\documentclass{article}

\usepackage{microtype}
\usepackage{graphicx} 
\usepackage{subfigure}
\usepackage{booktabs} %
\usepackage{comment}

\usepackage[utf8]{inputenc}
\usepackage{amsmath,amsthm,amssymb}
\usepackage{comment}
\usepackage{hyperref}
\usepackage[capitalize]{cleveref}
\usepackage{xcolor,xspace}
\usepackage{multirow}

\DeclareMathOperator*{\argmax}{\arg\,\max}

\newcommand{\ourmethod}{\textsc{LaMPP}\xspace}

\usepackage[arxiv]{icml2023}

\icmltitlerunning{LaMPP: Language Models as Probabilistic Priors for Perception and Action}

\usepackage[]{todonotes}

\makeatletter
\def\mathcolor#1#{\@mathcolor{#1}}
\def\@mathcolor#1#2#3{%
  \protect\leavevmode
  \begingroup
    \color#1{#2}#3%
  \endgroup
}
\makeatother

\begin{document}

\twocolumn[
\icmltitle{\ourmethod: Language Models as Probabilistic Priors for Perception and Action}

\begin{icmlauthorlist}
\icmlauthor{Belinda Z. Li}{mit}
\icmlauthor{William Chen}{mit}
\icmlauthor{Pratyusha Sharma}{mit}
\icmlauthor{Jacob Andreas}{mit}
\end{icmlauthorlist}

\icmlaffiliation{mit}{MIT CSAIL, Cambridge, Massachusetts, USA}

\icmlcorrespondingauthor{Belinda Z. Li}{bzl@mit.edu}

\icmlkeywords{Machine Learning, Natural Language Processing, Language Models, Priors, ICML}

\vskip 0.3in
]

\printAffiliationsAndNotice{}  %

\begin{abstract}

Language models trained on large text corpora encode rich distributional information about real-world environments and action sequences. This information plays a crucial role in current approaches to language processing tasks like question answering and instruction generation. We describe how to leverage language models for \emph{non-linguistic} perception and control tasks. Our approach casts labeling and decision-making as inference in probabilistic graphical models in which language models parameterize prior distributions over labels, decisions and parameters, making it possible to integrate uncertain observations and incomplete background knowledge in a principled way. Applied to semantic segmentation, household navigation, and activity recognition tasks, this approach improves predictions on rare, out-of-distribution, and structurally novel inputs.

\end{abstract}

\section{Introduction}
\label{sec:intro}

\begin{figure}[ht]
    \centering
    \includegraphics[width=\columnwidth,trim={0 14cm 36cm 0},clip]{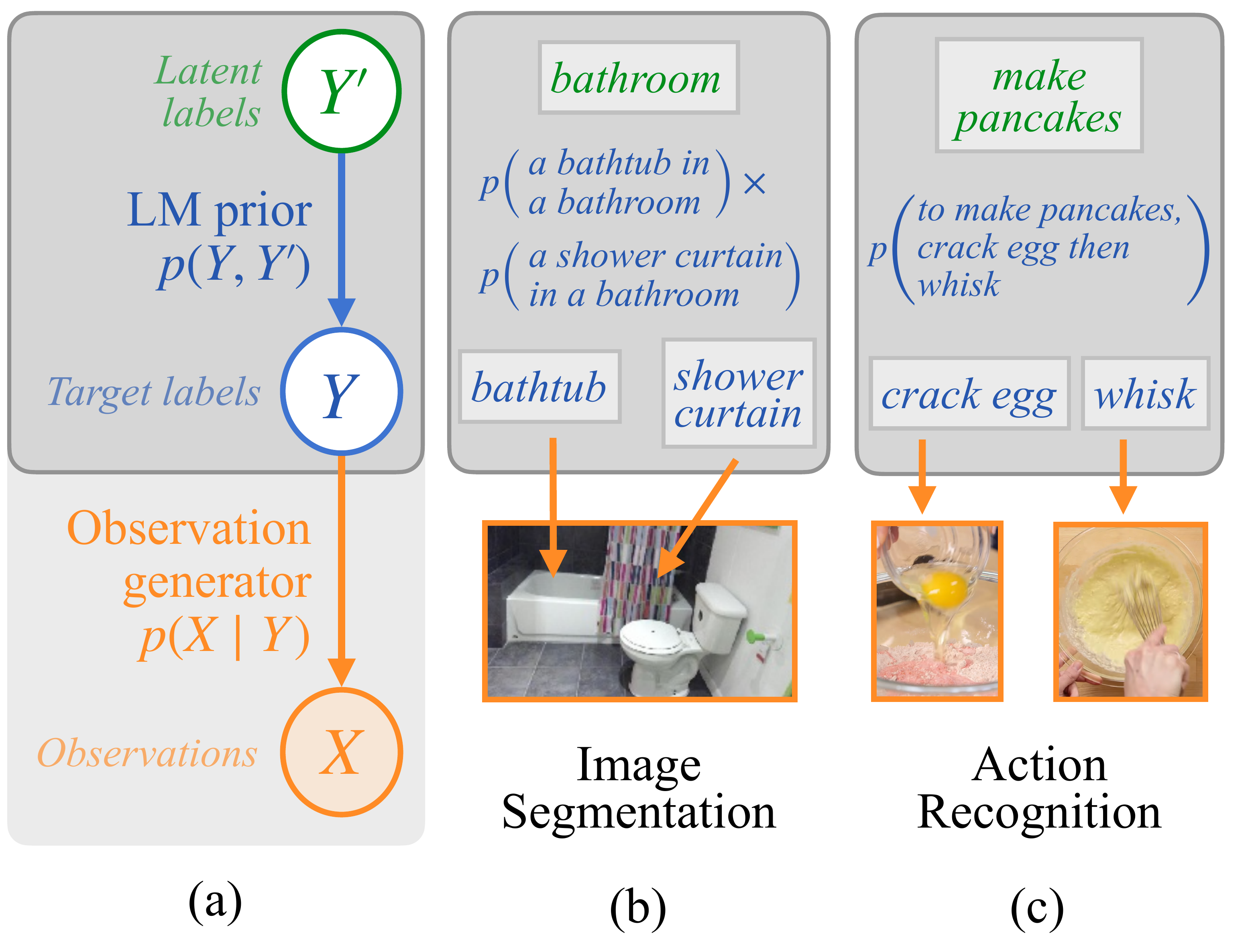}
    \vspace{-2em}
    \caption{
    In \ourmethod, the LM provides a prior over a structured label space $P(Y,Y')$ and a task-specific observation model provides $P(X\mid Y)$. We apply \ourmethod to three concrete tasks, including image segmentation %
    and video action recognition.*
    In the image segmentation case, the LM provides a prior over what objects are likely to co-occur (based on room-object probabilities), which allows it to determine that the observed curtain is a \textit{shower curtain}.
    In the action recognition case, the LM provides a prior over what action sequences are likely to accomplish the target tasks, allowing it to infer the action sequence in a video.
    \\
    \scriptsize{*Our third task, object navigation, is not shown in this figure.}
    }
    \vspace{-1em}
    \label{fig:teaser}
\end{figure}

\textbf{Common-sense priors} are crucial for decision-making under uncertainty in real-world environments.
Suppose that we wish to label the objects in the scene depicted in \cref{fig:teaser}(b). Once a few prominent objects (like the bathtub) have been identified, it is clear that the picture depicts a bathroom. This helps resolve some more challenging object labels: the curtain in the scene is a shower curtain, not a window curtain; the object on the wall is a mirror, not a picture.
Prior knowledge about likely object or event co-occurrences are essential not just in vision tasks, but also for navigating unfamiliar places and understanding other agents' behaviors.
Indeed, such expectations play a key role in human reasoning for tasks like object classification and written text interpretation~\cite{Kveraga2007-topDownPredictions, Mirault18-readWrong}.

In most problem domains, current machine learning models acquire information about the prior distribution of labels and decisions from task-specific datasets. %
Especially when training data is sparse or biased, this can result in systematic errors, particularly on unusual or out-of-distribution inputs. %
How might we endow models with more general and flexible prior knowledge?

We propose to use \textbf{language models}---learned distributions over natural language strings---as task-general probabilistic priors.
Unlike segmented images or robot demonstrations, large text corpora are readily available and describe almost all facets of human experience. Language models (LMs) trained on them encode much of this information---like the fact that \emph{plates are located in kitchens and dining rooms}, and that \emph{whisking eggs is preceded by breaking them}---often with greater diversity and fidelity than can be provided by small, task-specific datasets.
Such linguistic supervision has also been hypothesized to play a role in aspects of human common-sense knowledge that are difficult to learn from direct experience
\cite{Painter2005-learningThroughLanguage}.

In language processing and other text generation tasks, LMs have been used as sources of prior knowledge for tasks spanning common-sense question answering \cite{Talmor22-commonsenseQA2}, modeling scripts and stories \cite{Ammanabrolu21aaai-automatedStorytelling, Ammanabrolu21naacl-motivateDragon}, and synthesis of probabilistic programs \cite{Lew20-languagePPL}.
They have also been applied to grounded language understanding problems via \textbf{model chaining} (MC) approaches, which encode the output of perceptual systems as natural language strings that prompt LMs to directly generate labels or plans \cite{Zeng22arxiv-socraticmodels, Singh22-progprompt}.

In this paper, we instead focus on LMs as a source of probabilistic background knowledge that can be integrated with existing domain models. 
LMs pair naturally with structured probabilistic modeling frameworks: by using them to place \emph{prior} distributions over labels, decisions or model parameters, we can combine them with domain-specific generative models or likelihood functions 
to integrate ``top-down'' background knowledge with ``bottom-up'' task-specific predictors.
This approach offers a principled way to integrate linguistic supervision with structured uncertainty about non-linguistic variables, making it possible to leverage LMs' \emph{knowledge} even in complex tasks where LMs struggle with \emph{inference}.

We call this approach to modeling \textbf{\ourmethod} (\textbf{La}nguage \textbf{M}odels as \textbf{P}robabilistic \textbf{P}riors).
\ourmethod is flexible and applicable to a wide variety of problems.
We present three case studies featuring tasks with diverse objectives and input modalities---semantic image segmentation, robot navigation, and video action segmentation. \ourmethod consistently improves performance on rare, out-of-distribution, and structurally novel inputs, and sometimes even improves accuracy on examples within the domain model's training distribution.
These results show that language is a useful source of background knowledge for general decision-making, and that uncertainty in this background knowledge can be effectively integrated with uncertainty in non-linguistic problem domains.

\section{Method}
\label{sec:methods}
A language model (LM) is a distribution over natural language strings. LMs trained on sufficiently large text datasets become good models not just of grammatical phenomena, but various kinds of world knowledge
~\cite{Talmor22-commonsenseQA2,li-etal-2021-implicit}. Our work proposes a method for extracting probabilistic common-sense priors from language models, which can then be used to supplement and inform \emph{arbitrary} task-specific models operating over multiple modalities. These priors can be leveraged at multiple stages in the machine learning pipeline:

\textbf{Prediction:}\space\space\space In many learning problems, our ultimate goal is to model a distribution $p(y \mid x)$ over labels or decisions $y$ given (non-linguistic) observations $x$. These $y$s might be structured objects: in \cref{fig:teaser}(b), $x$ is an image and $y$ is a set of labels for objects in the image. 
By Bayes' rule, we can write:
\begin{equation}
p(y \mid x) \propto p(y) p(x \mid y) ~ ,
\end{equation} which factors this decision-making problem into two parts: a prior over labels $p(y)$,
and a generative model of observations $p(x \mid y)$. If we have such a generative model, we may immediately combine it with a representation of the prior $p(y)$ to model the distribution over labels.

\textbf{Learning:}\space\space\space In models with interpretable parameters, we may also leverage knowledge about the distribution of these parameters themselves during learning, before we make any predictions at all. Given a dataset $\mathcal{D}$ of examples $(x_i, y_i)$ and a predictive model $p(y \mid x; \theta)$, we may write 
\begin{align}
p(\theta \mid \mathcal{D}) &\propto p(\mathcal{D} \mid \theta) p(\theta) \nonumber \\  &= \Big(\prod_i p(y_i \mid x_i; \theta) \Big) p(\theta) ~ ,
\end{align}
in this case making it possible to leverage prior knowledge of $\theta$ itself, e.g., when optimizing model parameters or performing full Bayesian inference.

In structured output spaces, like segmented images, robot trajectories, or high-dimensional parameter vectors, a useful prior contains information about which joint configurations are plausible (e.g., an image might contain sofas and chairs, or showers and sinks, but not sinks and sofas).
How can we use an LM to obtain and use distributions $p(y)$ or $p(\theta)$? Applying \ourmethod in a given problem domain involves four steps:
\begin{enumerate}
\item \textbf{Choosing a base (domain) model}: Here we can use any model of observations $p(x \mid y)$ or labels $p(y \mid x; \theta)$.

\item \textbf{Designing a label space}: When reasoning about a joint distribution over labels or parameters, correlations between these variables might be expressed most compactly in terms of some other latent variable (in \cref{fig:teaser}(b), object labels are coupled by a latent \emph{room}). Before querying an LM to obtain $p(y)$ or $p(\theta)$, we may introduce additional variables like this one to better model probabilistic relationships among labels.

\item \textbf{Querying the LM}: We then obtain scores for each configuration of $y$ or $\theta$ by \emph{prompting} a language model with a query about the plausibility of the configuration, then \emph{evaluating} the probability that the LM assigns to the query. Examples are shown in \cref{fig:teaser}(b--c). For all experiments in this paper, we use the GPT-3 to score queries \cite{Brown20neurips-gpt3}.

\item \textbf{Inference}: Finally, we perform inference in the graphical model defined by $p(y)$ and $p(x \mid y)$ (or $p(\theta)\,p(y \mid x, \theta)$) to find the highest-scoring (or otherwise risk-minimizing) configuration of $y$ for a given $x$.
\end{enumerate}

In Sections~\ref{sec:img}--\ref{sec:video}, we apply this framework to three learning problems.
In each section, we evaluate \ourmethod's ability to improve \emph{generalization} over base models. We focus on three types of generalization:
\textbf{zero-shot} (ZS), \textbf{out-of-distribution} (OOD), and \textbf{in-distribution} (ID).
The type of generalization required depends on the availability and distribution of training data:
ZS evaluations focus on the case in which $p(x \mid y)$ is known (possibly just for components of $y$, e.g., appearances of individual objects), but no information about the joint distribution $p(y)$ (e.g., configurations of rooms) is available at training time.
OOD evaluations focus on biased training sets (in which particular label combinations are over- or under-represented). ID evaluations focus on cases where the full evaluation distribution is known and available at training time.

\section{\ourmethod for Semantic Segmentation}
\label{sec:img}

We first study the task of \textbf{semantic image segmentation}: identifying object boundaries in an image and labeling each object $x_i$  with its class $y_i$.
How might background knowledge from an LM help with this task? Intuitively, it may be hard for a bottom-up visual classifier to integrate global image context and model correlations among distant objects' labels. LMs encode common-sense information about the global structure of scenes, which can be combined with easy-to-predict object labels to help with more challenging predictions.

\subsection{Methods}
\begin{figure}[t]
    \centering
    \includegraphics[width=0.7\columnwidth,trim={0 8cm 41.5cm 0},clip]{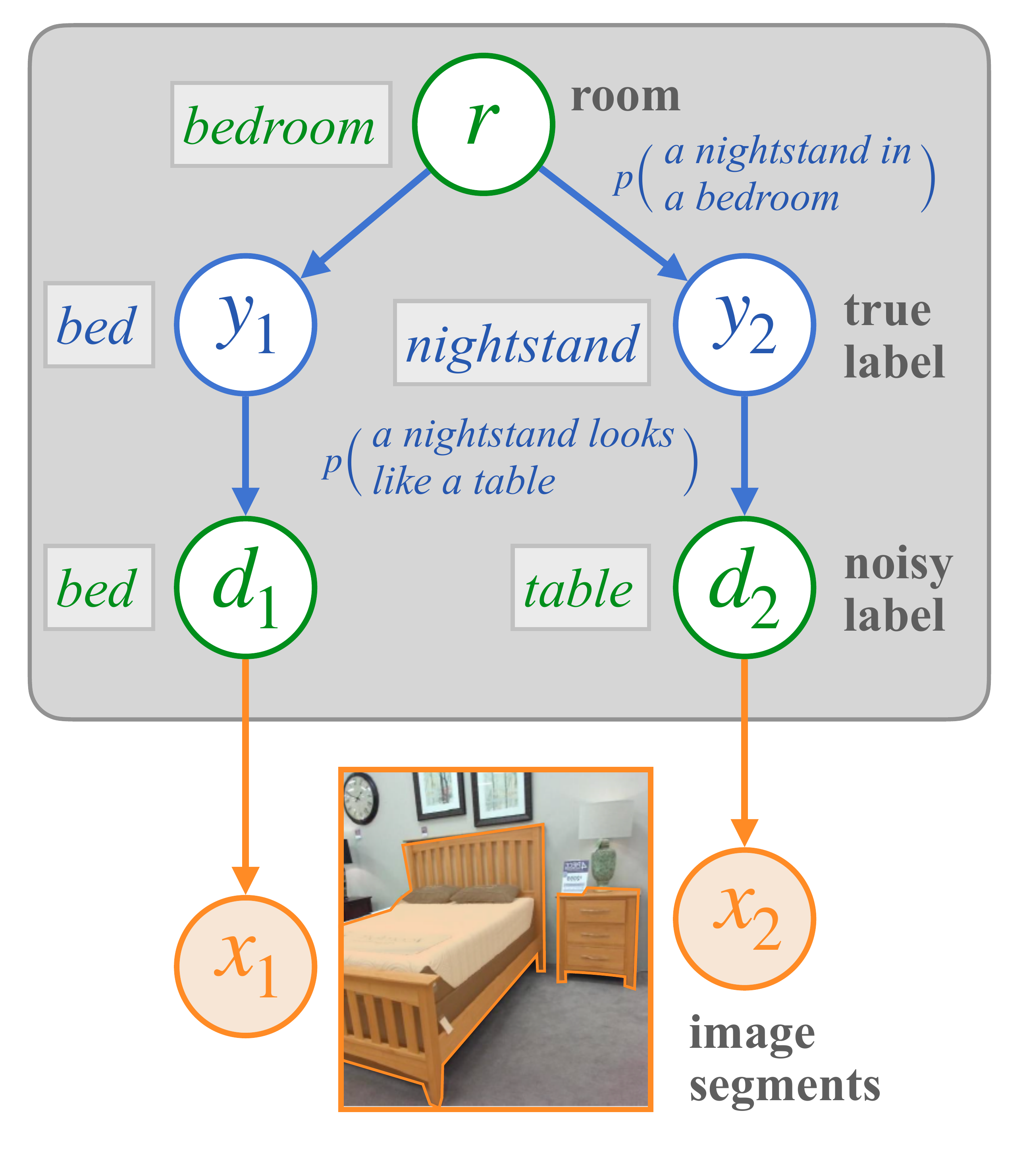}
    \vspace{-2em}
    \caption{Generative model for \textbf{image semantic segmentation}. Images originate in a room $r$, which generates the objects $y_1,y_2$ in the room, which generate noisy object labels $d_1,d_2$ representing perceptually similar objects. Finally, each $d_i$ generates an image segment $x_i$, a continuous region of image pixels depicting each object. Rooms $r$, true labels $y_i$, and noisy labels $d_i$ are latent, while image segments $x_i$ are observed.}
    \label{fig:image_seg}
\end{figure}

\paragraph{Base model}
Standard models for semantic segmentation discriminatively assign a label $y_i$ to each pixel $x_i$ in an input image $x$ according to some:
\begin{equation}
   \label{eq:img_seg_obj}
   p_\text{seg}(y_i \mid x) ~ .
\end{equation}
Our experiments use RedNet~\cite{rednet}, a ResNet-50-based autoencoder model, to compute \cref{eq:img_seg_obj}. By computing $\argmax_y ~ p(y \mid x)$ for each pixel in an input image, we obtain a collection of \textbf{segments}: contiguous input regions assigned the same label (see bottom of \cref{fig:image_seg}). When applying \ourmethod, we treat these segments as given, but attempt to choose a better joint labeling for all segments in an image.

\textbf{Label Space}\space\space\space
We do so using the generative model depicted in \cref{fig:image_seg}. We hypothesize a generative process in which every image originates in a \textbf{room} $r$. Conditioned on the room, a fixed number of \textbf{objects} are generated, each with label $y_i$. To model possible perceptual ambiguity, each true object labels in turn generates a \textbf{noisy} object label $d_i$. Finally, each of these generates an image segment $x_i$.

We use the base segmentation model $p_\text{seg}$ to compute $p(x_i \mid d_i)$ by applying Bayes' rule locally for each segment: $p(x_i \mid d_i) \propto p_\text{seg}(d_i \mid x_i) / p(d_i)$. 
All other distributions in this generative model are parameterized by an LM, as described below. Ultimately, we wish to recover ``true'' object labels $y_i$; the latent labels $r$ and $d$ help extract usable background information about objects' co-occurrence patterns and perceptual properties.

\textbf{LM Queries}\space\space\space We compute the object--room co-occurrence probabilities $p(y_i \mid r)$ by prompting the LM with the string:
\begin{center}
{\it A}({\it n}) [$r$] {\it has a}({\it n}) [$y_i$]: \colorbox{yellow}{[{\it plausible / implausible}]}
\end{center}
The LM conditions on the non-highlighted portion of the prompt and is expected to generate one of the highlighted tokens.
We compute the relative probability the LM assigns to tokens \emph{plausible} and \emph{implausible}, %
then normalize these over all object labels $y$ to parameterize the final distribution. We use the same procedure to compute the object--object confusion model $p(d_i \mid y_i)$, prompting the LM with:
\begin{center}
{\it The} [$d_i$] {\it looks like the} [$y_i$]: \colorbox{yellow}{[{\it plausible / implausible}]}
\end{center}

\textbf{Inference}\space\space\space
The model in \cref{fig:image_seg} defines a joint distribution over all labels $\underline{y} = y_1, \ldots, y_n$. To re-label a segmented image, we compute the max-marginal-probability label for each segment independently:
\begin{align}
&\argmax p(y_i \mid \underline{x}) \nonumber\\
&\quad = 
\argmax \sum_{r} \sum_{\underline{y} \setminus \{y_i\}} \sum_{\underline{d}} p(\underline{x}, \underline{d}, \underline{y}, r)
\end{align}
The form of the decision rule used for semantic segmentation (which includes several simplifications for computational efficiency) can be found in~\cref{sec:app_img_methods}. 

\subsection{Experiments}

We use the SUN RGB-D dataset for our semantic segmentation tasks \cite{Song15cvpr-sunrgbd}, which %
contains %
RGB-D images of indoor environments. %
We also implement a model-chaining (MC) baseline 
that integrates LM knowledge without considering model uncertainties.
We take noisy labels from the image model ($d_i$) and directly query the LM for true labels ($y_i$).
Details of this baseline can be found in~\cref{sec:app_img_MC}.
We attempt to make the MC inference procedure as analogous to our approach as possible: the LM must account for both room-object co-occurrence likelihoods %
and object-object resemblance likelihoods when predicting true labels. 
However, here, the LM must implicitly incorporate these likelihoods into its text-scoring, rather than integrating them into a structured probabilistic framework.
We evaluate the RedNet \textit{base model}, this \textit{model chaining} approach, and \textit{\ourmethod} on \textbf{in-distribution}
 and \textbf{out-of-distribution} generalization.
 
\textbf{ID Generalization}\space\space\space
We use a RedNet checkpoint trained on the entire SUNRGB-D training split. %
As these splits were not created with any special biases in mind, the training split should reflect 
a similar label distribution to the test split.

\textbf{OOD Generalization}\space\space\space
We study the setting where the training distribution's $p(y_i,y_j)$ differs from the true distribution's.
We do this by %
picking two object labels that commonly occur together (i.e.\ picking $y_i$ and $y_j$ such that $p(y_i,y_j)$ is high), and removing all images from the training set where they \textit{do} occur together (thus making $p(y_i,y_j)$ close to zero in the training set).
In particular, we choose bed and nightstand as these two objects, and 
hold out all images in the training set where nightstands and beds co-occur (keeping all other images). %
After training on this set, we evaluate on the original test split where beds and nightstands frequently co-occur.

\subsection{Results}
We evaluate the mean intersection-of-union (mIoU) between predicted and ground-truth object segmentations over all object categories for ID and OOD in \cref{tab:image_seg_results}.

\begin{table}[t]
    \centering
    \footnotesize
    \resizebox{\columnwidth}{!}{
    \begin{tabular}{llclll}
    \toprule
    	& Model & %
         mIoU &  Best/Worst Object ($\Delta$IoU) 
         \\
    \midrule
         \multirow{3}{*}{ID} & Base model & %
        47.8 & - \\
        & Model chaining & %
        37.5
        & shower curtain $(+16.9)$ \\
        & & & toilet $(-37.2)$ \\
        \\
        & \multirow{1}{*}{\ourmethod}
        & \multirow{1}{*}{48.3}
        & shower curtain $(+18.9)$ \\
        & & & desk ($-2.16$) \\
    \midrule
         \multirow{2}{*}{OOD} & Base model & %
        33.8 & - \\
        & \multirow{1}{*}{\ourmethod}
        & \multirow{1}{*}{34.0}
        & nightstand $(+8.92)$ \\
        & & & sofa ($-2.50$) \\
    \bottomrule
    \end{tabular}
    }
    \caption{Image semantic segmentation results for ID and OOD generalization. 
    We report Intersection-over-Union (IoU) for each model: the base model, a model chaining approach, and \ourmethod.
    We report mIoU (IoUs averaged over each object category), as well as the most- and least-improved object from each method relative to the base model (and the corresponding $\Delta$ IoU).
    \ourmethod improves semantic segmentation dramatically on certain categories, while having minimal effect on all other categories. 
    }
    \label{tab:image_seg_results}
\end{table}

In each setting, we compare the base model against \ourmethod.
We see that in both the 
ID and OOD cases, \ourmethod improves upon the baseline image model.
The improvements seem small in an absolute sense because we average over 37 object categories. To get a better understanding of the distribution of improvements over object categories, we report \textit{per-category differences in IoU} of our model \textit{relative} to the baseline image model.
The rightmost column of~\cref{tab:image_seg_results} shows the most-improved and least-improved object categories (and the corresponding IoU change for those categories).
We see in both settings that the top object category improved significantly while %
all other object categories were not significantly affected.

In the ID setting, the accuracy of detecting \textit{shower curtains} improves by nearly 20 points with \ourmethod, as the base model obtains near-0\% mIoU on shower curtains, almost always mistaking them for \textit{(window) curtains}. Here, background knowledge from language fixes a major (and previously undescribed) prediction error for a rare class.
In the OOD setting, the base image model sees far fewer examples of nightstands and consequently never predicts nightstands on the test data. (\textit{Nightstands} are frequently predicted to be \textit{tables} and \textit{cabinets} instead). 
This is likewise rectified with \ourmethod: background knowledge from language reduces model sensitivity to a systematic bias in dataset construction.

Finally, we see that the model chaining approach repairs prediction errors on the same rare class as \ourmethod in the ID setting, but it also introduces new prediction errors on far more classes.

\section{\ourmethod for Navigation}
\label{sec:navigation}

We next turn to the problem of \textbf{object navigation}. Here, we wish to build an agent that, given a goal object $g$ (e.g., a television or a bed), can take actions $a$ to explore and find $g$ in an environment, while using noisy partial observations $x$ from a camera for object recognition and decision-making.
Prior knowledge about where goal objects are likely located can guide this exploration, steering agents away from regions of the environment unlikely to accomplish the agent's goals.

\subsection{Methods}
\label{sec:nav_methods}
\begin{figure}[t]
    \centering
    \includegraphics[width=0.7\columnwidth,trim={0 12cm 40cm 0},clip]{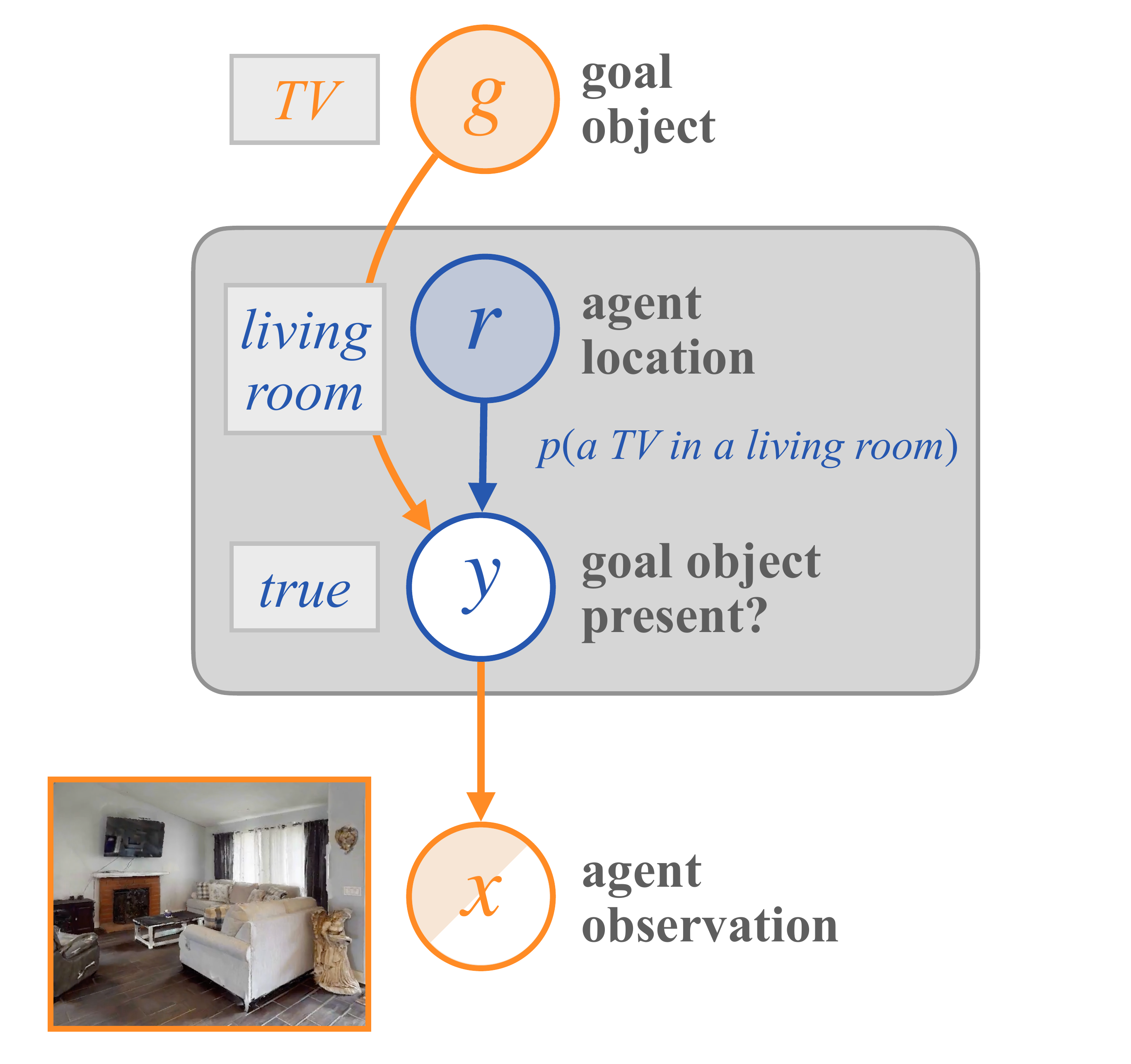}
    \vspace{-1em}
    \caption{Generative model for \textbf{object navigation}. Specifically, given a goal object $g$, we depict a decomposition of the agent's \textit{success score} $y$, which takes on value $1$ (\textit{true}) if the goal object is present and $0$ (\textit{false}) otherwise.
    We focus on a household domain, where any particular agent location must be within some room $r$. $r$ then generates the success condition $y$ (indicating whether $g$ is present at the agent location), which generates the agent observation $x$ of the location. Goal objects $g$ and rooms $r$ are given, success conditions $y$ are latent, and agent observations $x$ are partially-observed.}
    \label{fig:navigation}
\end{figure}

\textbf{Base model}\space\space\space We assume access to a pre-trained navigation policy (in this case, from the \textsc{stubborn} agent; \citealp{Luo2022-stubborn}) that can plan a path to any specified coordinate $a$ in the environment given image observations $x$. %
Our goal is to
build a \emph{high-level} policy $\pi(a \mid x)$
that can direct this low-level navigation.
We focus on navigation in household environments, and assume access to a coarse semantic map of an environment that identifies rooms, but not locations of objects within them.
In each state, the \textsc{stubborn} low-level navigation policy also outputs a scalar score reflecting its confidence that the goal object is present. 

\textbf{Label space}\space\space\space
Our high-level policy alternates between performing two kinds of actions $a$:
\begin{itemize}
  \item \textbf{Navigation}: the agent chooses a room $r$ in the environment to move to. (When a room is selected, we direct the low-level navigation policy to move to a point in the center of the room, and then explore randomly within the room for a fixed number of time steps.)
  
  \item \textbf{Selection}: whenever an observation is received \emph{during} navigation, the agent evaluates whether it has already reached the goal object. (When the goal object is judged to be present, the episode is ended.)
\end{itemize}

A rollout of this policy thus consists of a sequence of navigation actions, interleaved with a selection action for every observation obtained while navigating.
In both cases, choosing actions effectively requires inference of a specific unobserved property of environment state:
whether the goal object is in fact present near the agent. We represent this property with a latent variable $y$.
When navigating, the agent must infer the room that is most likely to contain the goal object. When selecting, the agent must infer whether its current perception is reliable.

We normalize the low-level policy's success score and interpret it as a distribution $p(x \mid y)$, then use the LM to define a distribution $p(y \mid r, g)$. Together, these give a distribution over latent success conditions and observations given goals and agent locations, which may be used to select actions in the high-level policy.

\textbf{LM queries}\space\space\space
For $p(y\mid r,g)$, we use the same query as in~\cref{sec:img} for deriving object--room probabilities, inserting $g$ in place of $y_i$,
except here 
we do not normalize over object labels (since $y$ is binary), and simply take the relative probability of generating the token \textit{plausible}.

\textbf{Inference}\space\space\space
With this model, we define a policy that 
performs inference about the location of the goal object, then
greedily attempts to navigate to the location most likely to contain it. This requires defining $p(a \mid x, g)$ for both navigation and selection steps.
\begin{itemize}
  \item \textbf{Navigation}: the agent chooses a room $r$ maximizing $p(y \mid r, g)$. (The agent does not yet have an observation from the new room, so the optimal policy moves to the room most likely to contain the goal object \emph{a priori}.)

  \item \textbf{Selection}: the agent ends the episode only if $p(y \mid x, r, g) > \tau$ for  some confidence threshold $\tau$.
\end{itemize}
During exploration, the agent maintains a list of previously visited rooms. Navigation steps choose only among rooms that have not yet been visited.

\subsection{Experiments}
\label{sec:navigation_task}
We consider a modified version of the Habitat Challenge ObjectNav task %
\cite{Yadav22-habitatchallenge}. 
The task objective is 
to find and move to an instance of the object in unfamiliar household environments as quickly as possible. 
The agent receives first-person RGBD images, compass readings, and 2D GPS values as inputs at each timestep.
In our version of the task, we assume access to a high-level map of the environment which specifies the coordinates and label of each room. Individual objects are not labeled; the agent must rely on top-down knowledge of where certain objects are likely to be in order to efficiently find the target object.

We implement a MC baseline where the LM guides agent exploration by specifying an ordering of rooms to visit.
This is similar to prior work that use LMs to specify high-level policies~\cite{Zeng22arxiv-socraticmodels,Sharma21-skillInduction}, 
whereby neither LM nor observation model uncertainties are accounted for when generating the \textit{high-level} policy. Details of the MC baseline can be found in~\cref{sec:app_nav_MC}.

We evaluate the ability of the original \textsc{stubborn} agent (\textit{base model}), \textit{model chaining}, and our agent (\textit{\ourmethod}) to perform \textbf{zero-shot generalization}, where the training data does not contain any information about $p(y\mid r,g)$.\footnote{At the time these experiments were conducted, room labels were not yet present in the dataset, so we could only study the zero-shot setting. To evaluate \ourmethod, the first two authors of the paper manually annotated room labels in the evaluation set.}
We also compare to a \textit{uniform prior baseline} where we preserve the high-level policy of our agent but replace LM priors over object-room co-occurrences %
with uniform priors:
\begin{align}
\begin{split}
    p(y\mid r,g) = \frac{1}{\text{\# room types in environment}}.
\end{split}
\end{align}
Note in the zero-shot case we have no additional information about $p(y\mid r,g)$, so we must assume it is uniform.

\subsection{Evaluation \& Results}
\label{sec:nav_results}
We evaluate success rate (SR), as measured by the percent of instances in which the agent successfully navigated to the goal object.
Because the \textsc{stubborn} agent is designed to handle only single floors (the mapping module only tracks a 2D map of the current floor),
we evaluate only instances in which the goal object is located on the same floor as the agent's starting location.

\begin{table}[t]
    \footnotesize
    \centering
    \begin{tabular}{llclcccccc}
    \toprule
        & \multicolumn{2}{c}{Success rate} &\\
        Model & Class & Freq. & Best/Worst Object ($\Delta$SR) \\
    \midrule
    Base model & 52.7 & 53.8 & - 
    \\
    Uniform prior & 52.1 & 51.7 & - 
    \\
    Model chaining & %
    61.2 & 65.3 & Toilet ($+20.9$) \\
    & & &  TV Monitor ($-4.2$)\\
    \\
    \ourmethod & 66.5 & 65.9 & TV Monitor ($+33.0$) \\
    & & & Plant ($-0.0$) \\
    \bottomrule
    \end{tabular}
    \caption{Navigation Results for ZS generalization. We report success rates (SR) for the base model, a uniform prior baseline model, a model chaining approach, and \ourmethod.
    We report both a \textit{class-averaged} SR (over goal objects) and a \textit{frequency-averaged} SR (over episodes). We also report the most-improved goal object and least-improved goal object for each method relative to the base model.
    We find that by using \ourmethod, we are able to achieve significant improvement over certain object classes. 
    }
    \label{tab:navigation_results}
\end{table}

Results are reported in~\cref{tab:navigation_results}. 
\ourmethod far outperforms both the base policy and the policy that assumes uniform priors, in overall and object-wise success rates. 
We find greatest improvements in
goal object categories that have strong tendencies to occur only in
specific rooms, such as TV monitors, and less for objects which tend to occur in many different rooms, like plants. 

Compared to the MC baseline, \ourmethod is better in terms of class-averaged SR, and comparable in terms of frequency-averaged SR. %
What accounts for this difference in performance?
In the MC approach, high-level decisions from the LM and low-level decisions from observation models are usually considered separately and delegated to different phases (it is hard to combine these information sources in string-space):
in our implementation, the MC baseline uses the top-down LM for \textit{navigation}, and the bottom-up observation model for \textit{selection}. %
Because the policy dictated by the \ourmethod probabilistic model also ignores bottom-up observation probabilities until the goal object is observed,
the \textit{navigation} step of both approaches is functionally equivalent. 
However, for the \textit{selection} step, 
we find that combining bottom-up and top-down uncertainties
is crucial;
in analyses in~\cref{sec:app_nav_analysis}, we see that when %
model uncertainties are ablated,
our agent actually \textit{underperforms} a comparable model chaining baseline.

Other than performance differences, \ourmethod is also substantially more query-efficient: MC requires one query per navigation action of \textit{each episode}, while \ourmethod simply requires a fixed number of queries ahead of time, which can be applied to \textit{all actions and episodes}.

\section{\ourmethod for Action Recognition and Segmentation}
\label{sec:video}

The final task we study focuses on video understanding: specifically, taking demonstrative videos of a task (e.g., making an omelet) and segmenting them into actions (e.g., cracking or whisking eggs).
Because it is hard to procure segmented and annotated videos, datasets for this task are usually small, and it may be difficult for models trained on task data alone to learn robust models of task-action relationships and action orderings. Large LMs' training data contains much more high-level information about tasks and steps that can be taken to complete them.

\subsection{Methods}
\begin{figure}[t]
    \centering
    \includegraphics[width=0.7\columnwidth,trim={0 4cm 38cm 1cm},clip]{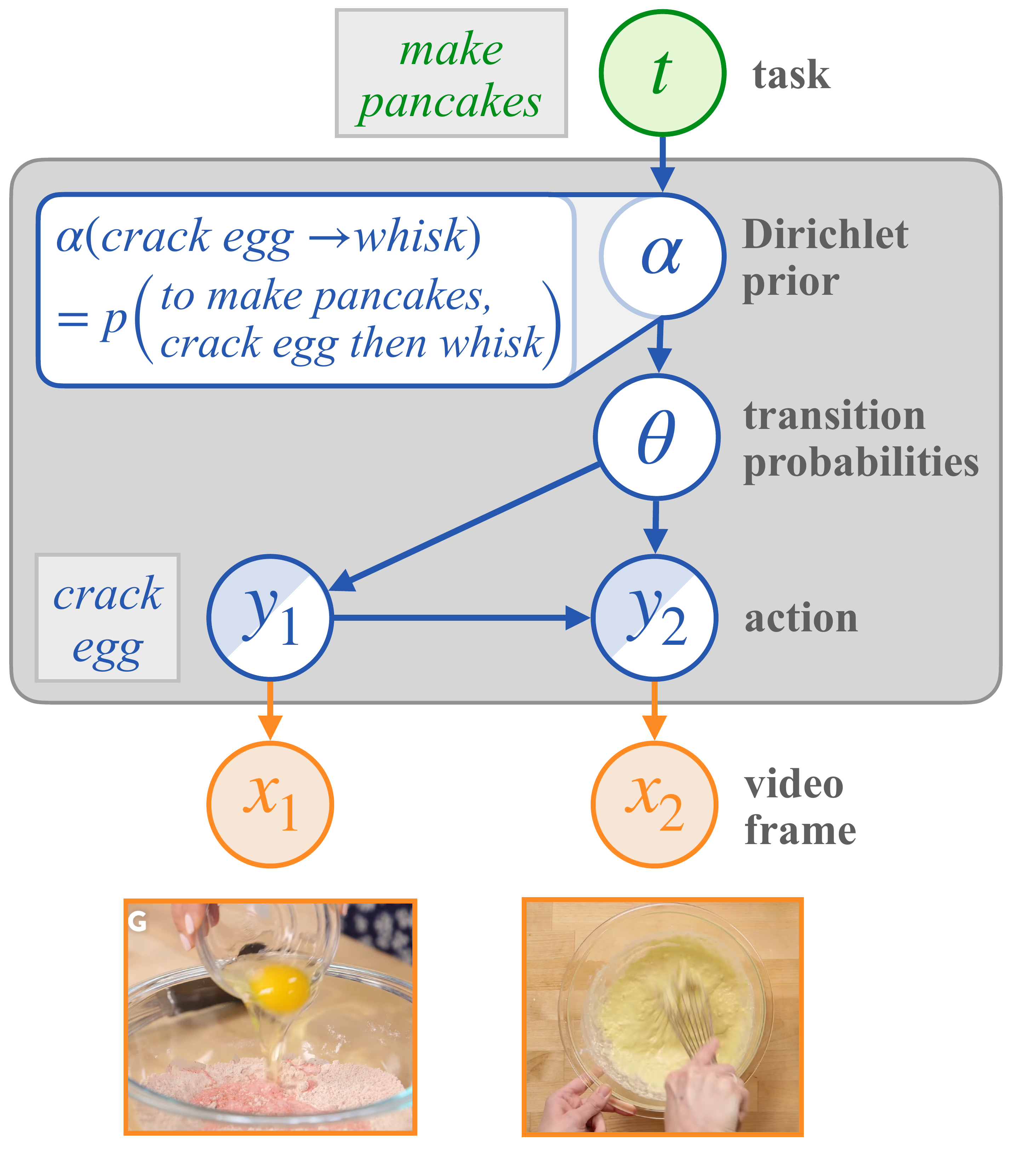}
    \caption{Generative model for \textbf{video-action segmentation}. 
    The base model we use for this task is a HMM with transition probabilities parameterized by $\theta$.
    In this task, we %
    generate a prior over \textit{model parameters} $\theta$:
    Each task $t$ generates a Dirichlet prior $\alpha$ over action transitions, which in turn generates model parameters $\theta$. $\theta$ parameterizes the action transition distribution $y_1\to y_2$. Each action $y_i$ at timestep $i$ then generates the observed video frame $x_i$. Tasks $t$ and video frames $x_i$ are observed, actions $y_i$ are partially-observed, and parameter priors $\alpha$ and parameters $\theta$ are latent.}
    \label{fig:vid_seg}
\end{figure}

\textbf{Base model}\space\space\space
Given a video of task $t$, we wish to label each video frame $x_i$ with an action $y_i$ (chosen from a fixed inventory of plausible actions for the task) according to:
\begin{align}
\label{eq:vid_act_seg_obj}
    \argmax_{y_1\cdots y_n} p(y_1\cdots y_n \mid x_1\cdots x_n, t).
\end{align}
We build on a model by \citet{fried-etal-2020-learning} that frames this as inference in a task-specific hidden Markov model (HMM) in which a latent sequence of actions generates a sequence of video frames according to a distribution: 
\begin{align}
&p(x_1, \ldots, x_n \mid y_1, \ldots, y_n) \nonumber \\
&\qquad \propto \prod_j p(x_j \mid y_j; \eta)\, p(y_j \mid y_{j-1}; \theta)
\end{align} 
(omitting the dependence on the task $t$ for clarity).
This generative model decomposes into an \textbf{emission model} with parameters $\eta$ and a \textbf{transition model} with parameters $\theta_t$, and allows efficient inference of $p(y \mid x)$.\footnote{\citet{fried-etal-2020-learning}'s model is a hidden \emph{semi}-Markov model (HSMM) in which latent action states generate multiple lower-level actions in sequence. While our experiments also use an HSMM, we omit the HSMM emission model for clarity of presentation.} $p(y_j \mid y_{j-1}; \theta)$ is a multinomial distribution parameterized by a table of transition probabilities, each of which encodes the probability that action $y_{j-1}$ is followed by action $y_j$.

In contrast to previous sections, which used pre-trained domain models, here we apply \ourmethod to the problem of learning model parameters themselves. Specifically, we use an LM to place a prior on \emph{transition parameters} $\theta$, making it possible to learn about valid action sequences from data while still incorporating prior knowledge from language. Given a dataset of labeled videos of the form $(x_{1...n}, y_{1...n})$, we compute a maximum \emph{a posteriori} estimate of $\theta$:
\begin{equation}
\label{eq:thetamap}
  \argmax_{\theta} ~ \log p(\theta) + \sum_{x, y} \sum_j \log p(y_j \mid y_{j-1}; \theta) ~ ,
\end{equation}
(likewise for $\eta$). At evaluation time, we use these parameter estimates to label new videos.

\textbf{Label space}\space\space\space
We parameterize the prior $p(\theta)$ as a Dirichlet distribution with hyperparameters $\alpha$, according to which:
\begin{equation}
p(\theta) \propto \prod_i \theta_i^{\alpha_i - 1} ~ .
\end{equation}
Intuitively, the larger $\alpha_i$ is, the more probable the corresponding $\theta_i$ is judged to be \emph{a priori}. Here, parameters $\theta_{y \to y'}$ are probabilities of transitioning from action $y$ to $y'$; we would like $\alpha_{y \to y'}$ to be large for plausible transitions, which is achieved by extracting values directly from a LM.

\textbf{Prompting the LM}\space\space\space
To derive values of $\alpha$ for each action transition $y \to y'$,
we query the LM with the prompt:
\begin{quote}
    \textit{Your task is to }[$t$]\textit{. Here is an *unordered* set of possible actions: \{}[$Y$]\textit{\}. Please order these actions for your task.} %
    \textit{The step after }[$y$]\textit{ can be }\colorbox{yellow}{[$y'$]}
\end{quote}
where $Y$ is a set of all available actions for the task. We condition the LM on the non-highlighted portion of the prompt and set $\alpha_{y \to y'} = \lambda \cdot p_\text{LM}(y' \mid \textrm{prompt}(y))$ (the probability of completing the prompt with the action name $y'$), where $\lambda$ controls the strength of the prior.

\textbf{Inference}\space\space\space
The use of a Dirichlet prior means that \cref{eq:thetamap} has a convenient closed-form solution:
\begin{align}
\theta_{y \to y'} =
\frac{
    \alpha_{y \to y'} + \#(y \to y') - 1
}{
    (\sum_{y''} \alpha_{y \to y''}) + \#(y) - |Y| 
} ~ ,
\end{align}
where $\#(y \to y')$ denotes the number of occurrences of the transition $y \to y'$ in the training data, $\#(y)$ denotes the number of occurrences of $y$ in the training data, and $|Y|$ is the total number of actions.

\subsection{Experiments}
We evaluate using the CrossTask dataset~\cite{crosstask}, which features instructional videos depicting tasks (e.g., \textit{make pancakes}). The learning problem is to segment videos into regions and annotate each region with the corresponding action being depicted (e.g., \textit{add egg}).

We evaluate the ability of %
the \textit{base model} and \textit{\ourmethod} to perform \textbf{zero-shot} and \textbf{out-of-distribution} generalization. For all experiments with \ourmethod, we use $\lambda = 10$. We do not study a MC baseline for this task, as model chaining is unable to generate parameters rather than labels.

\textbf{ZS Generalization}\space\space\space
We assume that the training data contains no information about the %
transition distribution $p(y_i\mid y_{i-1},t)$.
However, we still assume access to \textit{all video scenes and their action labels}, which allows us to learn emission distributions $p(x_i\mid y_i)$.
We do this by assuming access to only an \textit{unordered set} of video frames from each task, where each frame is annotated with its action label, but with no sense of which frame preceded or followed it.

Because we have no access to empirical counts of transitions from the training data,
the model falls back completely on its priors when computing 
 those parameters: %
    $$\theta_{y \rightarrow y'} = \frac{\alpha_{y \rightarrow y'} - 1}{(\sum_{y''} \alpha_{y \to y''}) - |Y|}$$
which is uniform for the base model %
and derived from the LM for \ourmethod.

\textbf{OOD Generalization}\space\space\space
We bias the \textit{transition distribution} by 
randomly sampling a common transition from each task and holding out %
all videos from the training set that contain that transition.

\subsection{Evaluation \& Results}
Following~\citet{fried-etal-2020-learning}, we evaluate \textit{step recall}, i.e. the percentage of actions in the real action sequence that are also in the model-predicted action sequence.
For simplicity, we ignore background actions during evaluation.

\begin{table}[t]
    \centering
    \footnotesize
    \begin{tabular}{llccccc}
    \toprule
        & & \multicolumn{2}{c}{Recall} \\
        & &  (class avg.) & 
         (freq avg.) 
        \\
    \midrule
        \multirow{2}{*}{ZS} & Base model & 44.4 & 46.0 
        \\
        & \ourmethod & 45.7 & 47.9 
        \\
    \midrule
        \multirow{2}{*}{OOD} & Base model & 37.6 & 40.9
        \\
        & \ourmethod &  38.1 & 41.2 
        \\
    \bottomrule
    \end{tabular}
    \caption{Video segmentation results for ZS and OOD generalization. We report step recall for the base model and \ourmethod. 
    We report both a \textit{class-averaged} step recall (over goal objects) and a \textit{frequency-averaged} step recall (over videos). We also report the most-improved action and least-improved action for \ourmethod relative to the base model in each setting.
    \ourmethod provides a significant improvement in certain task classes. 
    }
    \label{tab:vid_seg_results}
\end{table}

Results are shown in~\cref{tab:vid_seg_results}. 
For both the ZS and OOD settings, step recall slightly improves with \ourmethod.
The small magnitude of improvement may be because the LM sometimes does not possess a sensible prior over action sequences (compared to room--object co-occurrences, which it possesses accurate and calibrated priors for). 
For example, it is heavily biased towards returning actions in the order they are presented in the prompt.\footnote{We tried over 20 prompts, verifying whether the predicted action order looked sensible, but all yielded mixed results. We used the best prompt for these experiments.}

Indeed, the transitions that see most improvement with~\ourmethod 
are also the ones for which LM priors are more aligned with the test data than the training-set priors.
For example, in the OOD setting, the held-out transitions' recalls improve by an average of \textbf{8.2\%}.

\section{Related Work}

\textbf{String Space Model Chaining}\space\space\space
There has been much recent work in combining and composing the functionality of various models \textit{entirely in string space}. 
The Socratic models framework \cite{Zeng22arxiv-socraticmodels} proposes chaining together models operating over different modalities by converting outputs from each into natural language strings.  
Inter-model interactions are then performed purely in natural language.

While such methods have yielded good results in many tasks, like egocentric perception and robot manipulation \cite{Ahn22arxiv-sayCan}, they are fundamentally limited by the expressivity of the string-valued interface. 
Models often output useful features that cannot be easily expressed in language, such as graded or probabilistic uncertainty (e.g., in a traditional image classifier). Even if such information is written in string form, there is no guarantee that language models will correctly use it for formal symbolic reasoning-- today's largest LMs still struggle with arithmetic tasks expressed as string-valued prompts
\cite{Ye2022-lmUnreliability}. 

Concurrent to the present work is the approach of \citet{Choi22-lmPriors}, which similarly seeks to use language model scores as a source of common-sense information in other decision-making tasks. There, LMs are applied to feature selection, reward shaping, and casual inference tasks, rather than used to provide explicit priors for probabilistic models.

\textbf{LMs and Probabilistic Graphical Models}\space\space\space
Interpretation of LMs as composable probability distributions is well studied in pure language-processing tasks. Methods like chain-of-thought question-answering \cite{Wei22-chainOfThought}, thought verification \cite{Cobbe21-verifierMath}, and bootstrapped rationale-generation \cite{Zelikman22-star} may all be interpreted as probabilistic programs encoded as repeated language model queries \cite{Dohan22icml-lmCascades}. However, this analysis exclusively considers language tasks; to the best of our knowledge, the present work is the first to specifically connect language model evaluations to probabilistic graphical models in non-language domains.

{
}

\section{Conclusion}
We have described \ourmethod, a generic technique for integrating background knowledge from language into decision-making problems by extracting probabilistic \textit{priors} from language models.
\ourmethod improves zero-shot, out-of-distribution, and in-distribution generalization across image segmentation, household navigation, and video-action recognition tasks.
It enables principled composition of uncertain perception and noisy common-sense and domain priors, and
shows that language models' comparatively unstructured knowledge can be integrated naturally into structured probabilistic approaches for learning or inference.
The effectiveness of \ourmethod depends crucially on the quality of the LMs used to generate priors. While remarkably effective, today's LMs still
struggle to produce calibrated plausibility judgments for some rare tasks.
Improving LM knowledge representations is an important problem not just for \ourmethod but across natural language processing; as the quality of
LMs for core NLP tasks improves, we expect that their usefulness for \ourmethod will improve as well.

\section*{Acknowledgements}
This material is based upon work supported by the National
Science Foundation under Grant Nos. 2238240 and 2212310. BZL is supported by a NDSEG Fellowship.
We would like to thank Luca Carlone for valuable discussions regarding the design of navigation experiments.

\bibliography{main}
\bibliographystyle{icml2023}

\newpage
\appendix
\onecolumn %

\section{\ourmethod for Semantic Segmentation}
\subsection{Methods}
\label{sec:app_img_methods}

\newcommand{\appropto}{\mathrel{\vcenter{
  \offinterlineskip\halign{\hfil$##$\cr
    \propto\cr\noalign{\kern2pt}\sim\cr\noalign{\kern-2pt}}}}}
\newcommand{\obss}{\underline{x}}
\newcommand{\labs}{\underline{y}}
\newcommand{\olabs}{\underline{y} \setminus \{y_i\}}
\newcommand{\dets}{\underline{d}}

We derive the following decision rule from the model in \cref{fig:image_seg}:
\begin{align}
\label{eq:img_lm_app}
    p( & y_i \mid \underline{x}) \appropto \nonumber \\
    &
    p(y_i\mid d_i = d_i^*) p(d_i = d^*_i \mid x_i) \left( \sum_r p(r) p(y_i \mid r)  \prod_{j=1\cdots n}\left(  \sum_{y_j} \frac{p(r\mid y_j) p(d_j = y_j\mid x_j)}{p(r)}  \right)\right)
\end{align}

We obtain this decision rule as described below. Here
we denote rooms $r$, true object labels $y$, noisy object labels $d$, and %
observations $x$. (Underlines denote sets of variables, so e.g., $\obss = \{x_1, \ldots, x_n\}$.)
Finally, we
write $d_i^*$ to denote the base model's prediction for each image segment ($d_i^* = \arg\max p_\text{seg}(d_i\mid x_i)$). To see this:

\begin{align*}
    p(y_i \mid \obss) &\propto \sum_r \sum_{\olabs} \sum_{\dets} p(\obss, \labs, \dets, r) \\
    &= \sum_r \sum_{\olabs} \sum_{\dets} p(r) \Big( p(y_i \mid r) p(d_i \mid y_i) p(x_i \mid d_i) \Big) \prod_j p(y_j \mid r) p(d_j \mid y_j) p(x_j \mid d_j) \\
    &= \sum_r p(r) \Big( p(y_i \mid r) \sum_{d_i} p(d_i \mid y_i) p(x_i \mid d_i) \Big) \Big( \prod_j \sum_{y_j} p(y_j \mid r) \sum_{d_j} p(d_j \mid y_j) p(x_j \mid d_j) \Big) \\ 
    \intertext{
    Rather than marginalizing over all choices of $d$, we restrict each sum to a single term. For $d_i$, we choose the most likely detector output
    $d_i = d^*_i$. %
    For $d_j$, we choose the corresponding $y_j$ in the outer sum.
    Together, these simplifications reduce the total number of unnecessary LM queries about unlikely object confusions, and give a lower bound:
    }
    &\geq 
    p(d_i = d^*_i \mid y_i) p(x \mid d_i = d^*_i) 
    \sum_r p(r) (p(y_i \mid r) \Big( \prod_j \sum_{y_j}  p(y_j \mid r) p(d_j = y_j \mid y_j) p(x_j \mid y_j) \Big) \\
    \intertext{Applying Bayes' rule locally:}
    &= 
    \frac{p(y_i\mid d_i = d^*_i) p(d_i = d^*_i)}{p(y_i)} \frac{p(d_i = d^*_i \mid x_i) p(x_i)}{p(d_i = d^*_i)} \\
    &\qquad \sum_r p(r) p(y_i \mid r) \bigg( \prod_i \sum_{y_j} \frac{p(r \mid y_j) p(y_j)}{p(r)} p(d_j = y_j \mid y_j) \frac{p(d_j = y_j \mid x_j) p(x_j)}{p(d_j = y_j)} \bigg) \\
    \intertext{
    Finally, we make two modeling assumptions. First, we assume that of the form $p(y)$ and $p(d)$---the marginal distributions of true and noisy object labels---are uniform. This allows us to use LMs as a source of information about object co-occurrence probabilities without relying on their assumptions about base class frequency.
    Second,
    for non-target detections $x_j$, we assume the probability that 
    noisy labels match the true labels is constant over object categories.
    Then, dropping constant terms gives:}
    &\propto 
    p%
    (y_i\mid d_i = d^*_i) p%
    (d_i = d^*_i \mid x_i) 
    \sum_r p(r) p%
    (y_i \mid r) \Big( \prod_j \sum_{y_j} \frac{p%
    (r \mid y_j)}{p(r)} p%
    (d_j = y_j \mid x_j) \Big)
\end{align*}

\subsection{Model Chaining Baseline}
\label{sec:app_img_MC}
The model chaining baseline is given model predictions $\widehat{d_i}$ for \textit{each segment} $x_i$ and re-labels each segment by querying GPT-3 with:
\begin{quote}
\textit{You can see: }[$\underline{\widehat{d}}$] \\
\\
\textit{You are in the }\colorbox{yellow}{[$r$]} \\
\textit{The thing that looks like }[$\widehat{d_i}$]\textit{ is actually }\colorbox{yellow}{[$y_i$]}.
\end{quote}
The LM is given the non-highlighted portions and asked to generate the portions highlighted in yellow.
$\underline{\widehat{d}}$ is the set of all unique objects detected by the base model, written out as a comma-separated list.
$r$ is a room type generated by the LM based on these objects (inferred by normalizing over possible room types),
and $y_i$ is the actual identity of the object corresponding to this segment.
We replace all pixels formerly predicted as $\widehat{d_i}$ with $y_i$.

\section{\ourmethod for Navigation}
\label{sec:app_nav}

\begin{table}[t]
    \centering
    \begin{tabular}{lcc}
    \toprule
        Model & Class-Avg. SR & Freq.-Avg. SR \\
    \midrule
    \ourmethod & 66.5 & 65.9 \\
    $- p(y\mid r)$ during selection & 58.8 & 64.9 \\
    \midrule
    Model chaining & 61.2 & 65.3 \\
    \bottomrule
    \end{tabular}
    \caption{Navigation results verification ablations. We ablate the LM uncertainties over $p(y\mid r)$ when computing the selection action, making \ourmethod functionally similar to a model chaining baseline. We find that having these uncertainties are crucial; without them, \ourmethod actually \textit{underperforms} the model chaining baseline.}
    \label{tab:nav_results_ablations}
\end{table}

\subsection{Model Chaining Baseline}
\label{sec:app_nav_MC}
As in the image segmentation case, we have a model chaining baseline. LM priors are integrated into exploration through directly querying the LM with
\begin{quote}
\textit{The house has: }[$\underline{r}$]. \\
\textit{You want to find a }[$g$]\textit{. First, go to each }\colorbox{yellow}{[$r_0$]}\textit{. If not found, go to each }\colorbox{yellow}{[$r_1$]}\textit{. If not found, go to each $\cdots$ }
\end{quote}
whereby $\underline{r}$
is a list of all room types in the environment, for example, \textit{3 bathrooms, 1 living room, 1 bedroom}.
The LM returns the best room type $r_0$ to navigate to in order to find $g$. The agent visits all $r_0$ in order of proximity. If the object is not found, the LM is queried for the next best room type to visit, etc., until the object is found or we run out of rooms in the environment.

\subsection{Additional Analysis}
\label{sec:app_nav_analysis}
Why does \ourmethod outperform model chaining? As noted in~\cref{sec:nav_results}, model chaining approaches do not use bottom-up observational probabilities or top-down LM probabilities when generating their high-level policy. Our method does, specifically integrating both probabilities when performing selection (recall we threshold $p(y\mid x,r,g)$ at selection steps, which decomposes to $p(y\mid x)p(y\mid r,g)$). %
The model chaining equivalent to this phase simply delegates 
selection to the low-level model, which only uses observational uncertainties $p(y\mid x)$.

To further understand and how using LM probabilities contributes at this phase, we run a version of \ourmethod where we simply change the decision rule at the %
selection action to $p(y\mid x)$.
Results are reported in~\cref{tab:nav_results_ablations}.
Note that we actually \textit{underperform} the MC baseline when we take away top-down uncertainties $p(y\mid r,g)$ --- once again highlighting the importance of combining both sources of uncertainty.

\end{document}